\def\year{2020}\relax
\documentclass[letterpaper]{article} 
\usepackage{aaai20}  
\usepackage{times}  
\usepackage{helvet} 
\usepackage{courier}  
\usepackage[hyphens]{url}  
\usepackage{graphicx} 
\usepackage{graphicx}  
\usepackage{amsmath}
\usepackage{amssymb}
\usepackage{tabularx}
\usepackage{multirow}
\usepackage{multicol}
\usepackage{subcaption}
\usepackage{booktabs}

\title{Optimal Feature Transport for Cross-View Image Geo-Localization}

\author{Yujiao Shi,\textsuperscript{\rm 1, 2}
Xin Yu,\textsuperscript{\rm 1, 2}
Liu Liu,\textsuperscript{\rm 1, 2}
Tong Zhang,\textsuperscript{\rm 1, 3}
Hongdong Li\textsuperscript{\rm 1, 2}
\\
\textsuperscript{\rm 1}Australian National University, Canberra, Australia.\\
\textsuperscript{\rm 2}Australian Centre for Robotic Vision, Australia.\\
\textsuperscript{\rm 3}Motovis Australia Pty Ltd\\
\{firstname.lastname\}@anu.edu.au
}

\def\ie{\emph{i.e.}}
\def\eg{\emph{e.g.}}

\def\etal{\emph{et al.}}

\begin{document}

\maketitle

\begin{abstract}
This paper addresses the problem of cross-view image geo-localization, where the geographic location of a ground-level street-view query image is estimated by matching it against a large scale aerial map (\eg, a high-resolution satellite image).  State-of-the-art deep-learning based methods tackle this problem as deep metric learning which aims to learn global feature representations of the scene seen by the two different views.  Despite promising results are obtained by such deep metric learning methods, they, however, fail to exploit a crucial cue relevant for localization, namely, the spatial layout of local features.  Moreover, little attention is paid to the obvious domain gap (between aerial view and ground view) in the context of cross-view localization.   This paper proposes a novel {\em Cross-View Feature Transport (CVFT)} technique to explicitly establish cross-view domain transfer that facilitates feature alignment between ground and aerial images.   Specifically, we implement the CVFT as network layers, which transports features from one domain to the other, leading to more meaningful feature similarity comparison.  Our model is differentiable and can be learned end-to-end. 
Experiments on large-scale datasets have demonstrated that our method has remarkably boosted the state-of-the-art cross-view localization performance, \eg, on the CVUSA dataset, with significant improvements for top-1 recall from 40.79\% to 61.43\%, and for top-10 from 76.36\%  to 90.49\%.  
We expect the key insight of the paper (\ie, explicitly handling domain difference via domain transport) will prove to be useful for other similar problems in computer vision as well. 
\end{abstract}

\begin{figure}[!ht] 
    \centering
    \includegraphics[width=\columnwidth]{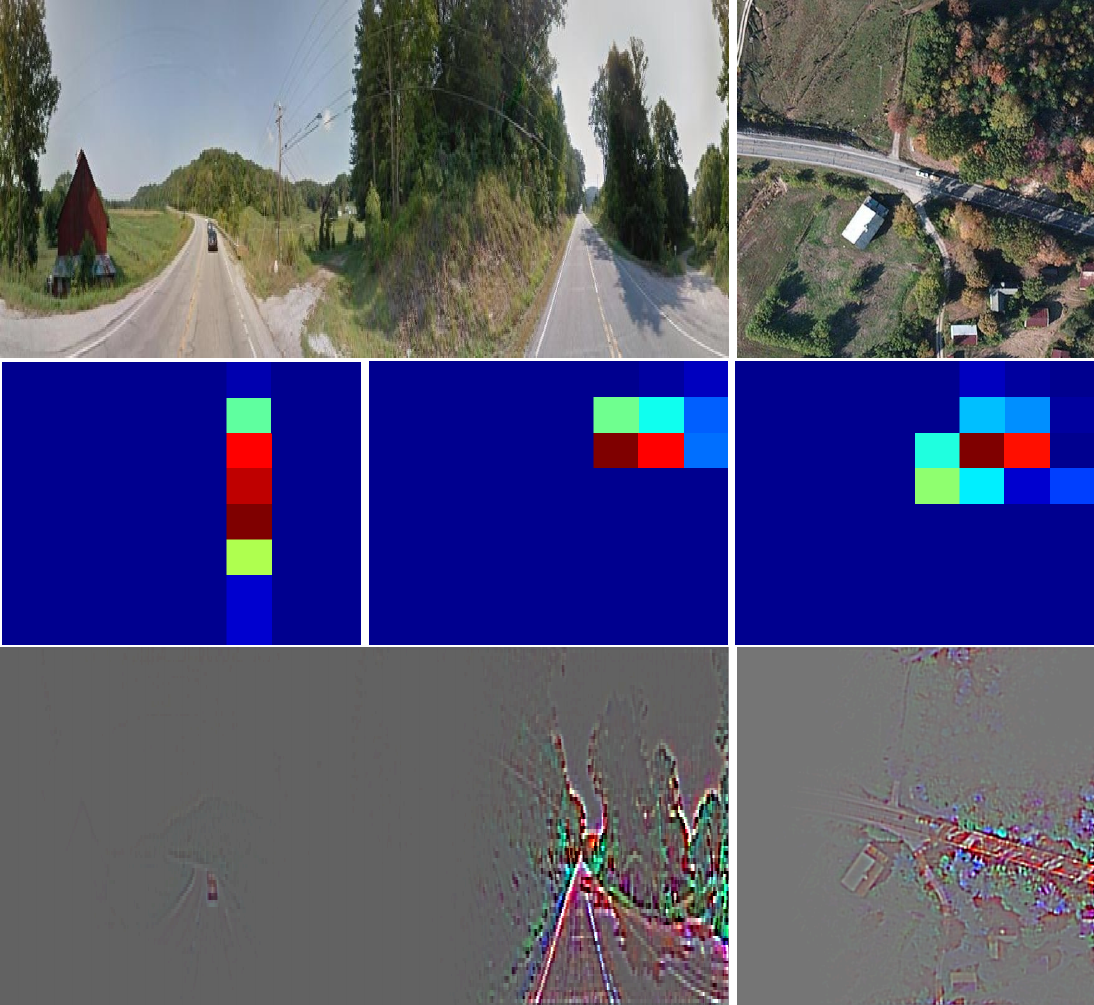}
    \caption{Illustration of the proposed Cross-View Feature Transport (CVFT). First row: a panoramic ground-level query image (left), and its corresponding aerial image (right); Second row: left: an example feature map learned from the ground image shown above; middle: the learned feature map from the aerial image shown above; Right: the transported ground-level feature map using our CVFT. It is clear that CVFT effectively aligns the two feature domains, bringing the two corresponding feature maps closer in the feature space; Bottom row: visualizing the two feature maps \cite{zeiler2014visualizing}.}
    \label{fig:correspondence}
\end{figure}

\section{Introduction}

This paper studies the problem of ground-to-aerial cross-view image-based localization. The task is: given a ground-level image of street scene (\eg, a Google street view image), we want to estimate where this image was taken, in terms of estimating its geo-location relative to a large (and geo-tagged) satellite map covering the region of interest (\eg, an entire city).  Such a cross-view localization problem is not only practically relevant, thanks to the wide-accessibility of high-resolution satellite imagery, but also scientifically important and challenging.  The challenges mainly come from the significant differences in their different viewpoints (one is from overhead, and the other at ground level) and vast variations in visual appearance.  

In this paper, we propose an effective mechanism to explicitly take into account the two issues: (1) preserving local feature spatial layout information, and (2) implementing a mechanism to account for feature domain differences. We first employ a two-branch Siamese-like CNN architecture to extract features from ground and aerial images separately. As seen in Figure \ref{fig:correspondence}, there is a spatial domain correspondence between ground panorama and aerial imagery.
In order to exploit the spatial information of local features, we propose a Cross-View Feature Transport (CVFT) module to explicitly transform feature maps from one domain (ground) to the other one (aerial). 

Specifically, we re-arrange feature maps extracted from one branch by a transport matrix. 
Then the transformed features from one domain will lie much closer to the corresponding features in the other domain in terms of positions, facilitating similarity comparisons. 
Moreover, the spatial layout information is also embedded in the features, which makes our features more discriminative. 

Our transport matrix is achieved by optimizing a cross-view feature transport problem using differentiable Sinkhorn operation. 
Since all the layers in our network are differentiable, 
our CVFT module can be trained in an end-to-end fashion. Extensive experimental results demonstrate that our CVFT improves cross-view geo-localization performance by a large margin on standard benchmarks.

The contributions of this work are as follows:

\begin{itemize}
\item A new state-of-the-art cross-view image-based geo-localization method, which outperforms all previous best performing deep networks developed for this task. 
\item A Cross-View Feature Transport (CVFT) module for domain transfer. Our CVFT module explicitly establishes feature transformations between two different domains, facilitating cross-domain feature similarity matching.
\end{itemize}

\section{Related works}
\noindent \textbf{Cross-view image-based Geo-localization:}
Benefited from recent advance of deep learning, most of the existing works on cross-view image matching adopt two-branch structured CNNs to learn representations for ground and aerial images and treat this task as a deep metric learning problem.
The seminal work \cite{workman2015location} fine-tuned AlexNet \cite{NIPS2012_4824} on Imagenet \cite{russakovsky2015imagenet} and Places \cite{zhou2014learning}, and then applied the network to extract features for cross-view matching/localization. 
Workman \etal \cite{workman2015wide} further indicated that fine-tuning the aerial branch by minimizing the distance between aerial and ground images leads to better localization performance. 
Later, Vo and Hays \cite{vo2016localizing} conducted thorough experiments on the investigation of the formulation and solutions towards this problem, \ie, binary classification or image retrieval, Siamese or Triplet network architecture.
Building upon the Siamese network, Hu \etal \cite{Hu_2018_CVPR} embedded a NetVLAD layer on top of a VGG network to extract descriptors that are invariant against large viewpoint changes.
Liu \& Li \cite{Liu_2019_CVPR} 
incorporated per-pixel orientation information into a CNN to learn orientation-selective features for cross-view localization.
Though effective these methods are, they all dismiss the domain difference and discard the spatial layout information in ground and aerial images. 

\noindent \textbf{Optimal Transport:} 
The optimal transport (OT) theory provides a principled way of comparing distributions and offers an optimal transport plan for matching distributions. 
However, as the classic OT is a Linear Programming problem, it suffers from expensive computation complexity in terms of (distribution or data) dimensions.
Regarding this deficiency, Cuturi \etal \cite{cuturi2013sinkhorn} proposed an entropy regularized optimal transport problem which can be solved by the Sinkhorn-Knopp algorithm  \cite{sinkhorn1967concerning,knight2008sinkhorn} efficiently.  
After that, it has been successfully applied to solve various machine learning problems, such as domain adaptation \cite{courty2017optimal}, multi-label learning \cite{frogner2015learning}, feature dimensionality reduction \cite{gautheron2018feature}, generative models \cite{park2018representing} and so on.
Inspired by the optimal transform, we construct a feature transport module in this paper to bridge the domain gap and encode the correspondence between ground and aerial features, and the Sinkhorn-Knopp algorithm is employed as a solver to the feature transport problem. 

\section{Motivation: Cross-View Feature Transport}
\begin{figure}[t!]
    \centering
    \begin{subfigure}[b]{0.9\columnwidth}
        \centering
        \includegraphics[width=\columnwidth]{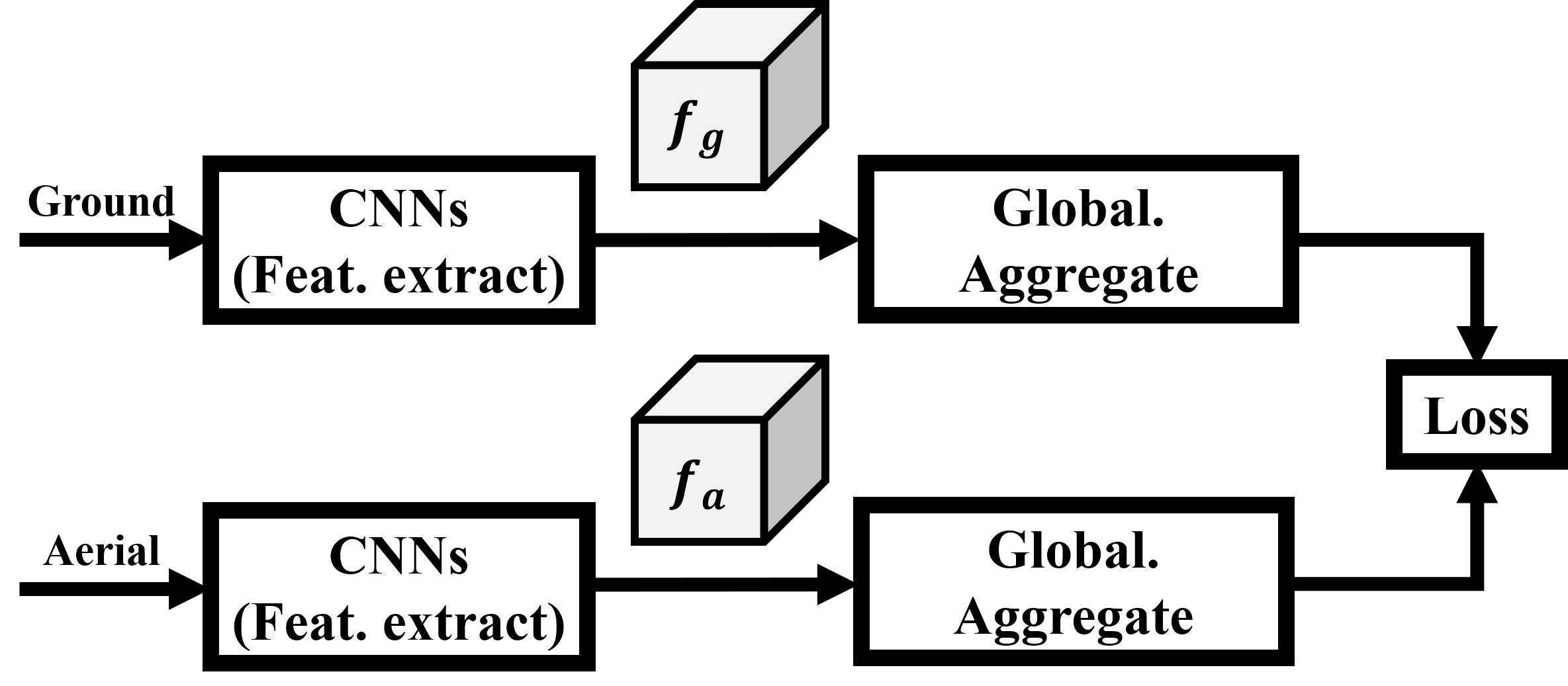}
        \caption{Previous architecture.}
        \label{fig:others_work}
    \end{subfigure}\\
     \begin{subfigure}[b]{0.9\columnwidth}
        \centering
        \includegraphics[width=\columnwidth]{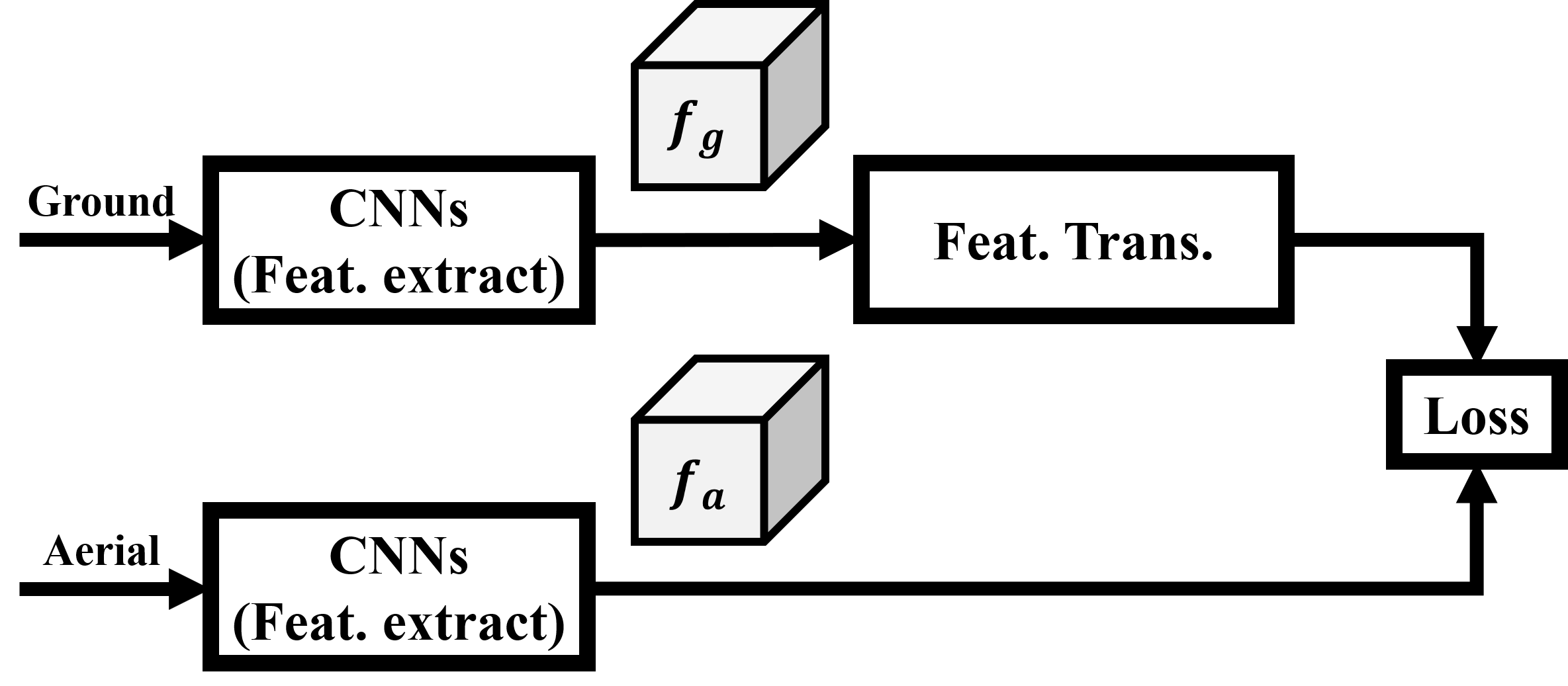}
        \caption{Our architecture.}
          \label{fig:Ours_framework}
    \end{subfigure}
    \caption{Comparison between traditional methods and ours. (a) Existing deep metric learning network with a global feature aggregation layer;(b) Our new network with Feature Transport.  Global Feature Aggregation tends to discard very useful spatial layout information. In contrast, our Feature Transport maps feature from the ground-view domain to the aerial domain, accounting for their domain difference.  Details of the Feature Transport will be explained in the main text.}
  \label{fig:structure_comprison}
\end{figure}

Existing deep learning based cross-view localization methods often adopt a forced feature embedding procedure that brings matching ground-and-aerial image pairs closer in the feature embedding space while far apart for those non-matching ones.
For instance,
a Global Aggregation Operation (such as `global pooling' or VLAD-type feature aggregation \cite{arandjelovic2016netvlad}) is often adopted.  

However, these global aggregation steps often discard (or destroy) the spatial layout information of local deep features (in other words, the relative positions between those local features in the image plane). This spatial layout information is however very useful and plays a pivotal role in aiding human's geo-localization (\eg, re-orientation or way-finding).  For example, human usually memorizes the relative spatial position of a certain object or building in a scene for the localization and navigation purpose. 
Thus it is important to exploit such cues for the image-based geo-localization. 

Besides, there are obvious domain differences between a ground-view image and a satellite image (namely,  the former is a panoramic image seen from the ground-level, while the latter depicts an overhead (bird-eye's) view of a geographic region). Hence, simply computing the distance between two deep-features from different domains may neglect the important domain correspondences, thus yielding inferior estimation.

In this paper, we take a direct (and explicit) approach to tackle this issue to account for the domain gap, such as drastic appearance changes as well as geometric distortions. Inspired by the idea of Optimal Transport (OT), we aim to resort a transport matrix to achieve this goal. Specifically, we propose a novel Feature Transport block ( in the form of differentiable neural layers) which is able to transport (transfrom or map) features from one domain (\eg, ground) to another (\eg, aerial).   Figure \ref{fig:structure_comprison} shows a comparison between the traditional method using global feature aggregation, and our new method with Feature Transport.  

To speed up the computation and make the feature transport objective can be optimized by a deep neural network, we employ an entropy-regularized optimal feature transport which can be solved by the differentiable Sinkhorn operation. 
The achieved feature transport plan is a doubly stochastic matrix encoding the relationships between cross-domain features. It can be used to transport ground features into the aerial domain such that matching ground and aerial features can be spatially aligned. 
In the next section, we will give more technical details about the process.

\begin{figure*}[!th]
    \centering
    \includegraphics[width=1.8\columnwidth]{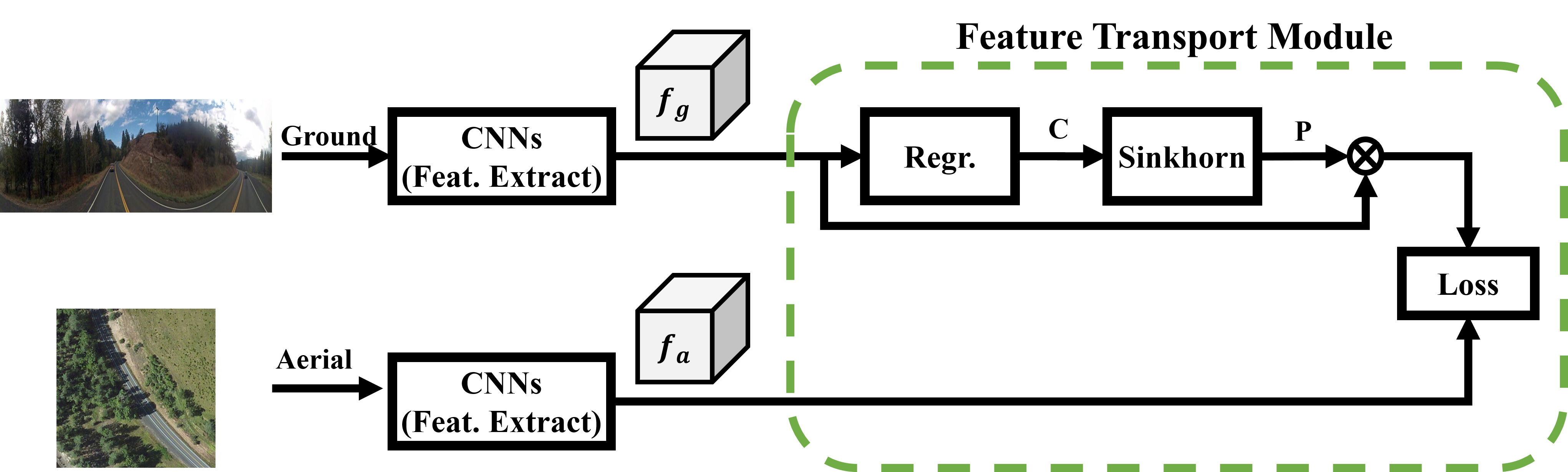}
    \caption{The framework of our method. A series of convolutional layers are applied to function as feature extractors, and then the extracted features are fed into a feature transport module to construct the correspondence between ground and aerial domains. }
    \label{fig:overall_framework}
\end{figure*}

\section{The Proposed CVFT Framework}

Figure \ref{fig:overall_framework} illustrates the overview of our proposed CVFT framework. 
We first employ a two-branch Siamese-like architecture to learn feature maps for ground and aerial images separately. 
In order to align the ground-view features to aerial-view features, we then exploit the idea of optimal transport and construct a Feature Transport Module to formulate the feature alignment. 
To be specific, we design a cost generation sub-network to formulate the feature transport problem, and a Sinkhorn solver to optimize this problem and obtain the feature transport matrix. 
Then, the feature transport matrix is applied to the ground-view feature maps. 
In this way, we re-arrange the ground-view feature maps and thus align them to their corresponding aerial ones as well as bridge the domain gap.
Our neural network is optimized by a metric learning objective in an end-to-end fashion. 
Detailed illustrations of each block in the framework will be presented in the following content.

\subsection{Feature Extraction}
Due to the powerful feature representation ability of deep convolutional neural networks, we employ a CNN, such as VGG \cite{simonyan2014very}, as our backbone network to extract features from input images. 
Since our backbone CNN, VGG network, outputs the feature maps with a high dimension (\ie, $8\times8\times512$), we apply another convolutional layer to reduce the feature dimension along the channel dimension rather than the spatial dimension. In this manner, we preserve the spatial layout information of the extracted features. Our final output feature dimension is $8\times8\times64=4096$, which is a commonly used feature representation dimension for image retrieval/matching task \cite{Hu_2018_CVPR}. 

\subsection{Optimal Feature Transport}
Optimal Transport (OT) theory was originally developed for finding the optimal transport plan $\mathbf{P}^*$ that can best align two probability distributions $\mathbf{\mu}_s$ and $\mathbf{\mu}_g$. This is done by minimizing the following transport distance:$\left \langle \mathbf{P}, \mathbf{C} \right\rangle_{F}$, where $\left \langle \cdot, \cdot \right \rangle_{F}$ is the Frobenius norm between two matrices, $\mathbf{C} \in \mathcal{R}^{n_s \times n_t}$ is a cost matrix measuring cost for transporting between source and target samples. $\mathcal{P}$ is a set of transport plans and $\mathbf{P} \in \mathcal{R}^{n_s \times n_t}$.  $n_s$ is the number of source samples $\mathbf{x}_s$ and $n_t$ is the number of target samples $\mathbf{x}_t$.

In the cross-view image-based geo-localization, we aim to expose the relationship (\ie,{feature transport plan}) between ground and aerial features so as to facilitate the cross-domain feature alignment. With this purpose, we generalize the optimal transport from distributions to features. 

\subsubsection{Cost Matrix Generation:}
In the cross-view feature transport, we first need to define the cost matrix between the ground features $\mathbf{f}^{i}(g) \in \mathcal{R}^{h\times w}$ and aerial features $\mathbf{f}^{i}(a)\in \mathcal{R}^{h\times w}$, where $h$, $w$ indicate the height and width of feature maps\footnote{For simplicity, we resize the ground and aerial feature maps into the same size before they are fed into the feature transport module. The source code of this paper is available at https://github.com/shiyujiao/cross\_view\_localization\_CVFT.git.}, and $i$ is the index of feature channels $c$. 

Ideally, both the ground and aerial images should be used to compute the cost matrix $\mathbf{C}$.  However, it is computationally expensive to calculate the cost matrix for each gallery image in large-scale retrieval problems. 
To speed up the efficiency of calculating $\mathbf{C}$, we propose to use one image from one domain to learn how to transform into the other domain. The other domain information is learned from the final objective function and the cost-matrix generation network is updated through backpropagation. As seen in Figure \ref{fig:overall_framework}, we employ a regression block to generate a cost matrix $\mathbf{C}$.

\subsubsection{Sinkhorn Solver:}

As the original OT is a Linear Programming problem that requires formidable computational costs regarding to the data dimension, we employ an entropy regularized version \cite{cuturi2013sinkhorn} which formulated the feature transport problem as a strictly convex problem that can be solved by the Sinkhorn operation efficiently:
\begin{small}
\begin{equation}\label{Eq::OT_minimization}
    \mathbf{P}^* = \mathop{\arg\min}_{\mathbf{P} \in \mathcal{P}} \left \langle \mathbf{P}, \mathbf{C} \right \rangle_{F} - \lambda h(\mathbf{P}),
\end{equation}
\end{small}where $h(\mathbf{P})$ is an entropy regularization term of $\mathbf{P}$ and $\lambda$ is a trade-off weight. The entropy regularization term forces the solution of Equation~\eqref{Eq::OT_minimization} to be smoother as $\lambda$ increases, and sparser as $\lambda$ decreases. $\mathbf{P}^*$ is a doubly stochastic matrix.

Given the cost matrix $\mathbf{C}$, the feature transport objective, as presented in Equation~\eqref{Eq::OT_minimization}, is constructed. We then adopt the Sinkhorn-Knopp algorithm to optimize the objective function and generate feature transport plans for the cross-view features. 

Specifically, the Sinkhorn algorithm first applies an exponential kernel on the cost matrix $\mathbf{C}$, yielding $\mathbf{C}' = \exp(-\lambda\mathbf{C})$. Second, the Sinkhorn algorithm normalizes rows and columns iteratively in order to convert $\mathbf{C'}$ to a doubly stochastic matrix. Denote
row $\mathcal{N}^{r}(\cdot)$ and column $\mathcal{N}^{c}(\cdot)$ normalization as follows:

\begin{small}
\begin{equation}
        \mathcal{N}_{i,j}^r(\mathbf{C}')=\frac{c'_{i,j}}{\sum_{k=1}^{N}c'_{i,k}}, ~~~~ \mathcal{N}_{i,j}^c(\mathbf{C}')=\frac{c'_{i,j}}{\sum_{k=1}^{N}c'_{k,j}},
\end{equation}
\end{small}where $c'_{i,j}$ represents an element in $\mathbf{C}'$.

For the $m$-th iteration, the output of Sinkhorn operation $\mathcal S (\cdot)$ can be represented in a recursively manner:

\begin{small}
\begin{equation}
    \mathcal S^{m}(\mathbf{C}') = \left\{\begin{matrix}
    \mathbf{C}' & m=0,\\ 
    \mathcal{N}^c(\mathcal{N}^r(S^{m-1}(\mathbf{C}'))) & otherwise.
    \end{matrix}\right.
\end{equation}
\end{small}
When iterations converge, we thus obtain our feature transport matrix $\mathbf{P}^* = \mathcal S^{m}(\mathbf{C}')$.
The Sinkhorn operation is differentiable and its gradient with respect to the input can be calculated by unrolling the sequence of the row and column normalization operations \cite{cuturi2013sinkhorn,santa2017deeppermnet}.
For example, the partial derivatives when $m=1$ can be defined as follows:
\begin{small}
\begin{equation}
    \frac{\partial \mathcal{S}^1 }{\partial c'_{s,t}} = 
\frac{\partial \mathcal{N}^{c}_{s,t}}
{\partial \mathcal{N}^{r}_{s,t}}
\cdot 
\sum_{j=1}^{N}
\left[\frac{[j=t]_{\mathbb{I}}}{\sum_{k=1}^{N}c'_{s,k}}\!-\! \frac{c'_{s,j}}{(\sum_{k=1}^{N}c'_{s,k})^2} \right],
\end{equation}
\end{small}where $[\cdot]_{\mathbb{I}}$ represents an indication function, and $s, t$ and $j$ represent the indices of the row and columns in $\mathbf{C}'$.

After obtaining $\mathbf{P}^*$, we can transport the feature maps between the ground-view and aerial-view images as follows:
\begin{small}
\begin{equation}
        \mathbf{f}^{i}(a) = n_a  \mathbf{P}^*\mathbf{f}^{i}(g), ~~\mathbf{f}^{i}(g) = n_g  \mathbf{P}^{*T}\mathbf{f}^{i}(a), i = 1,\ldots, c,
\end{equation}
\end{small}where $n_a$ and $n_g$ represent the feature numbers of aerial and ground images in each channel, respectively.

\subsection{Triplet Loss on transported features}
After transporting the ground-view feature maps, the ground-view features will be aligned to the corresponding overhead-view features. 
Then, we apply a metric learning objective to learn feature embedding for both the aligned ground-view images as well as aerial-view images. 
The triplet loss is widely used as an objective function to train deep neural networks for image localization and matching tasks \cite{Hu_2018_CVPR,Liu_2019_CVPR,vo2016localizing}. The goal of the triplet loss is to bring matching pairs closer while pushing non-matching pairs far apart. Therefore, we employ a weighted soft-margin triplet loss, $\mathcal{L}$, as our objective to train our neural network, expressed as:

\begin{small}
\begin{equation}
    \mathcal{L}= \log(1+e^{\gamma (d_{pos} - d_{neg})}),
\end{equation}
\end{small}where $d_{pos}$ and $d_{neg}$ represent the $\ell_2$ distance of all the positive and negative aerial features to the anchor ground feature, 
$\gamma$ is a parameter which controls the convergence speed \cite{Hu_2018_CVPR}.

\section{Experiments}

\begin{figure}[!t]

    \centering
    \includegraphics[width=\columnwidth]{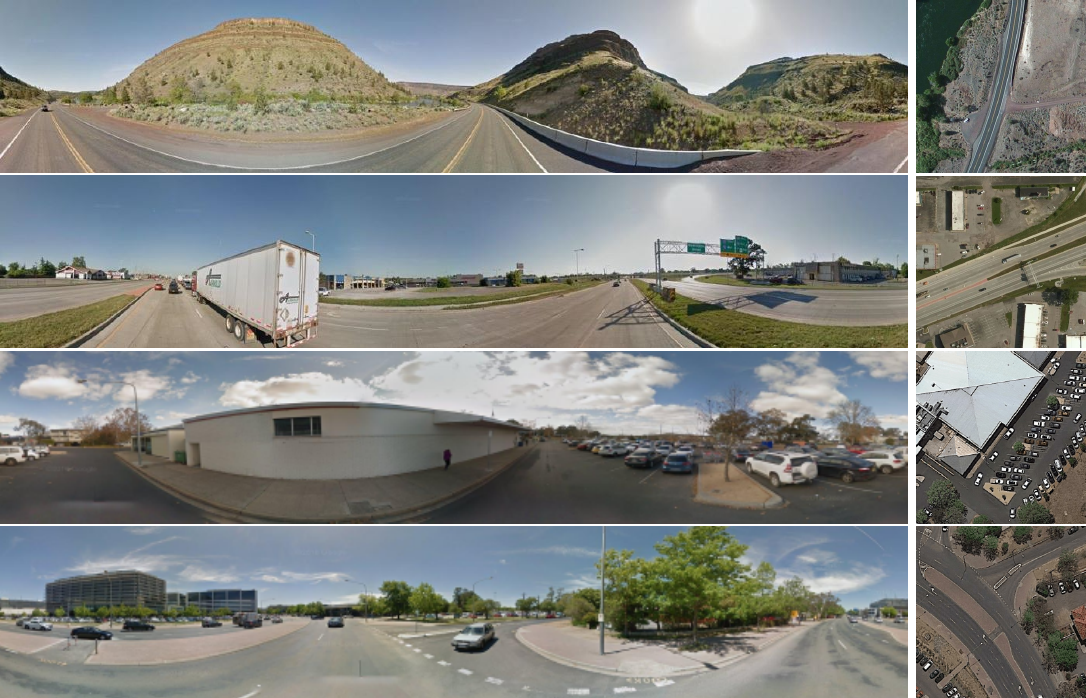}
    
    \caption{Samples of ground and aerial image pairs. Left are ground panorama images and right are the aerial images. The first two pairs are from CVUSA dataset and the second two pairs are from CVACT dataset.}
    
    \label{fig:CVUSA_ACT}
\end{figure}

\noindent \textbf{Training and Testing Datasets:}
We conduct our experiments on two standard benchmark datasets, namely CVUSA \cite{zhai2017predicting} and CVACT \cite{Liu_2019_CVPR}, for evaluation and comparisons. These two datasets are both cross-view datasets, and each contains 35,532 ground-and-aerial image pairs for training. CVUSA provides 8,884 image pairs for testing and CVACT provides the same number of pairs for validation (denoted as CVACT\_val). CVACT \cite{Liu_2019_CVPR} also provides 92,802 cross-view image pairs with accurate Geo-tags to evaluate Geo-localization performance (denoted as CVACT\_test). Figure \ref{fig:CVUSA_ACT} illustrates some sample pairs from these two datasets.

\noindent \textbf{Implementation Details:}
We employ VGG16 with pre-trained weights on ImageNet \cite{deng2009imagenet} as our backbone network to extract image features. The convolutional feature maps of the conv5\_3 layer in VGG16 are extracted as our image deep features. 
Similar to the work of \cite{Hu_2018_CVPR,Liu_2019_CVPR}, we set $\gamma$ to $10$ for the weighted soft-margin triplet loss.
Our network is trained using Adam optimizer \cite{kingma2014adam} with a learning rate of $10^{-5} $ and batch size of $B_s=12$. We exploit an exhaustive mini-batch strategy \cite{vo2016localizing} to construct the maximum number of triplets within each batch. For instance, for each ground-view image, there is $1$ positive satellite image and $B_s-1$ negative satellite images, and we can construct $B_s(B_s-1)$ triplets in total. Similarly, for each satellite image, there is $1$ positive ground-view image and $B_s-1$ negative ground-view images, and thus $B_s(B_s-1)$ triplets can also be established. 
Hence, we have $2B_s(B_s-1)$ triplets in total within each batch. 

\noindent \textbf{Evaluation Metrics:}
We apply the same evaluation metric as \cite{Hu_2018_CVPR,Liu_2019_CVPR,vo2016localizing}, known as r@K, to exam the performance of our method as well as compare with state-of-the-art cross-view localization methods. r@K measures the performance that how many satellite images in the database need to be retrieved to find the true matching image. More detailed explanations of r@K metric can be found in \cite{Hu_2018_CVPR,vo2016localizing}.  

\begin{table*}[!ht]
\centering
\caption{Recall performance on CVUSA \cite{zhai2017predicting} and CVACT\_val dataset \cite{Liu_2019_CVPR}.}
\label{tab::recall_compare_baseline}
\begin{tabular}{c|c|c|c|c|c|c|c|c}
\toprule
\multirow{2}{*} & \multicolumn{4}{c|}{CVUSA} & \multicolumn{4}{c}{CVACT\_val}  \\ \cline{2-9} 
                                  & r@1           & r@5            & r@10          & r@1\%     & r@1           & r@5           & r@10         & r@1\%   \\ \hline \hline
Liu \& Li \cite{Liu_2019_CVPR}     & 40.79         & 66.82          & 76.36         & 96.12         & 46.96         & 68.28         & 75.48         & 92.04 \\
CVM-NET \cite{Hu_2018_CVPR}       & 22.47         & 49.98          & 63.18         & 93.62         & 20.15         & 45.00         & 56.87         & 87.57 \\ 
VGG global pooling                & 31.53         & 59.85          & 70.91         & 95.09         & 28.98         & 57.04         & 67.96         & 91.72 \\ 
Our network wo/ CVFT              & 41.68         & 70.71          & 80.71         & 97.70         & 45.28         & 71.74         & 80.14         & 95.28 \\ 
Our network                       & \textbf{61.43}& \textbf{84.69} & \textbf{90.49}& \textbf{99.02}& \textbf{61.05}&\textbf{81.33} & \textbf{86.52}& \textbf{95.93}      \\ 
Our network (orientation noise =$\pm$20$^\circ$)& 52.07       & 76.87          & 84.58         & 97.77         & 54.81         & 77.70         & 83.37         & 94.92 \\ 
\bottomrule

\end{tabular}
\end{table*}

\subsection{Effects of Cross-view Feature Transport}

In this part, we conduct an ablation study to demonstrate the effectiveness of our proposed Cross-View Feature Transport (CVFT) module and how much our CVFT contributes to improving the geo-localization performance.

\subsubsection{Baseline Networks:} To illustrate the effectiveness of our proposed CVFT module, we simply remove it from our framework while keeping all the other parts unchanged as our first baseline, named Our network wo/ CVFT. Then we retrain this baseline. In this case, we still preserve the relative spatial information of local features but do not explicitly construct the cross-view correspondences in learning the network.
The performance of our first baseline is shown in the fourth row in Table \ref{tab::recall_compare_baseline}. The recalls for the top-1 candidate on CVUSA and CVACT\_val datasets are 41.68\% and 45.28\%, respectively. In comparison to our proposed network, we conclude that it is difficult for our first baseline to establish such spatial correspondences or transformations, thus leading to inferior localization performance.

Furthermore, in order to demonstrate the importance of the feature spatial layout information, we employ another baseline, called VGG global pooling. For this baseline, we apply global max pooling to the feature maps extracted by the layer of conv5\_3 in VGG16 as our image representation. Then we retrain this baseline. 
As seen in Table \ref{tab::recall_compare_baseline}, the recall performance for our second baseline is the worst. This phenomenon indicates that the feature spatial layout information plays a critical role in the cross-view geo-localization task.

\subsubsection{Our Network with CVFT:} 
\begin{figure}[!t]
    \centering
    \includegraphics[width=\columnwidth]{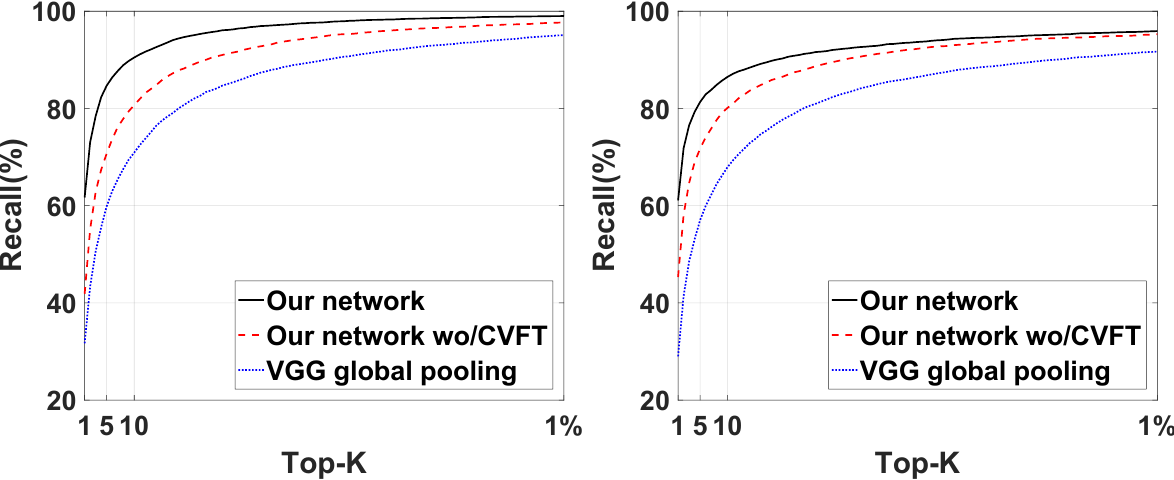}
    \caption{Comparison of Recalls on CVUSA (\textbf{Left}) and CVACT\_val (\textbf{Right}) datasets. Our method outperforms baseline methods by a large margin, with a relative improvement of $19.75\%$ and $15.77 \%$ at Top-1 for CVUSA and CVACT\_val datasets, respectively.}
    \label{fig:ourVSbaseline}
\end{figure}

Since our proposed network not only explores the spatial layout information but also explicitly establishes the domain correspondences via our CVFT, our network achieves the best performance as indicated by Table \ref{tab::recall_compare_baseline}. 
As visible in Figure \ref{fig:ourVSbaseline}, the performance of our network is consistently better than our baselines, and our network outperforms the baseline networks by a large margin on both CVUSA and CVACT\_val datasets.
For example, we obtain over \textbf{15\%} improvements on recall@1 on both datasets with the help of our CVFT module.

\subsection{Visualization of Feature Alignment}
\begin{figure}[!t]
\centering
    \begin{subfigure}[t]{\columnwidth}
    \centering
        \includegraphics[width=0.85\columnwidth]{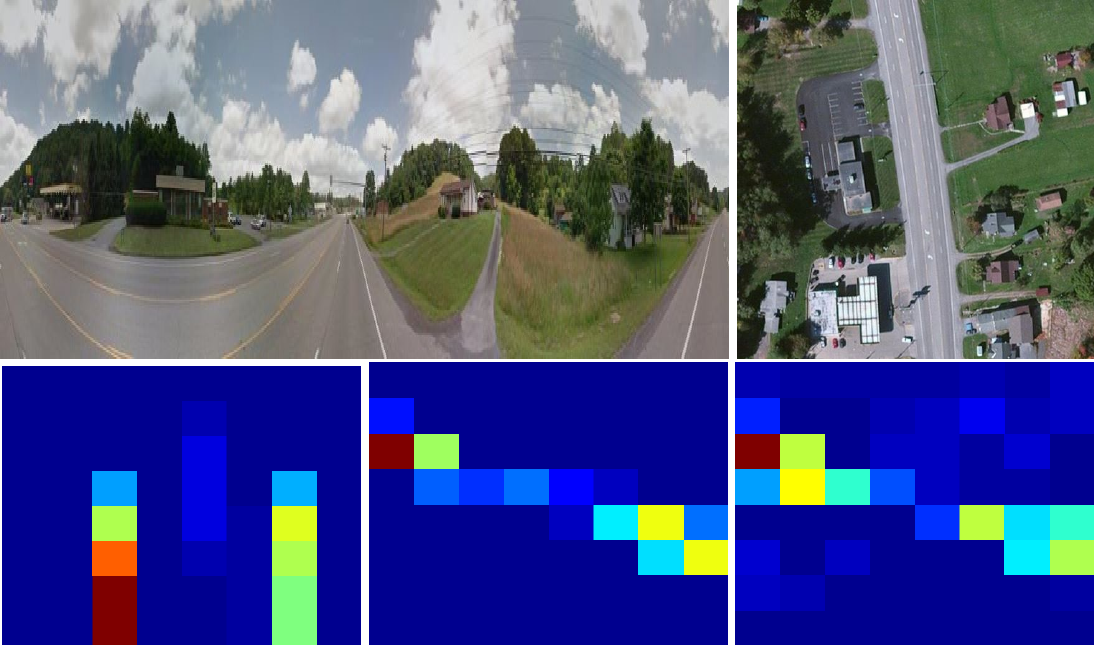}
        \caption{}
    \end{subfigure}\\
    \begin{subfigure}[t]{\columnwidth}
    \centering
        \includegraphics[width=0.85\columnwidth]{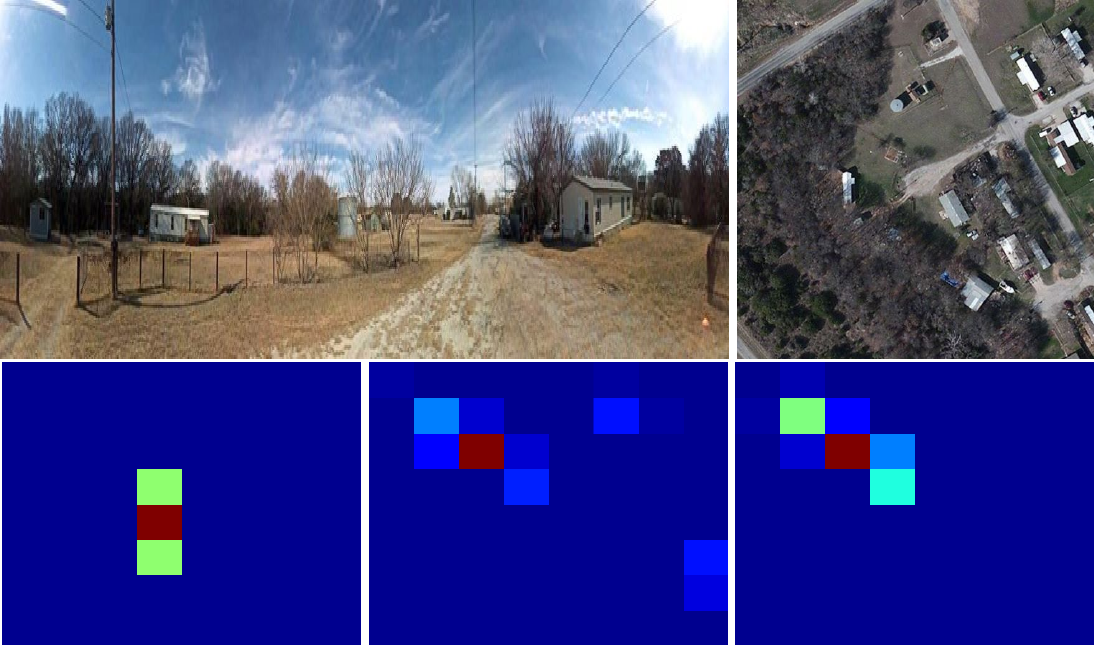}
        \caption{}
     \end{subfigure}
\caption{Our method successfully aligns cross-view feature maps by using CVFT feature transport. For the second row of each subfigure: (Left:) Input ground-view feature map; (Middle:) The corresponding satellite image feature map; (Right:) CVFT-transported ground feature map. Note the two feature maps are better aligned.}

\label{fig:CVFT_feature_map_alignment}
\end{figure}

The core insight of our method is that our CVFT is able to align the feature maps from a cross-view image pair. To verify this intuition, we show two examples of cross-view feature map alignments in Figure \ref{fig:CVFT_feature_map_alignment}.
As visible in Figure \ref{fig:CVFT_feature_map_alignment}, we demonstrate two features from the ground-view images (the first column) as well as two features from the aerial-view images (the second column). After applying our estimated feature transport plan $\mathbf{P}^*$ to the ground-view features, we obtain transported ground features (the third column). Compared with the features directly extracted from the aerial images, our transported ground features are well aligned to them. This experiment not only shows the effectiveness of our CVFT but also demonstrates that our network can explicitly establish the relationship between the ground and aerial domains, thus facilitating the cross-view matching.  
Note that our transport matrix $\mathbf{P}$ is different from spatial transformer networks (STN) \cite{jaderberg2015spatial}, since cross-view feature transformation cannot be modeled by geometric transformation of few parameters. Detailed illustration and experiments are presented in supplementary material.

\begin{figure}[!t]
    \centering
    \includegraphics[width=\columnwidth]{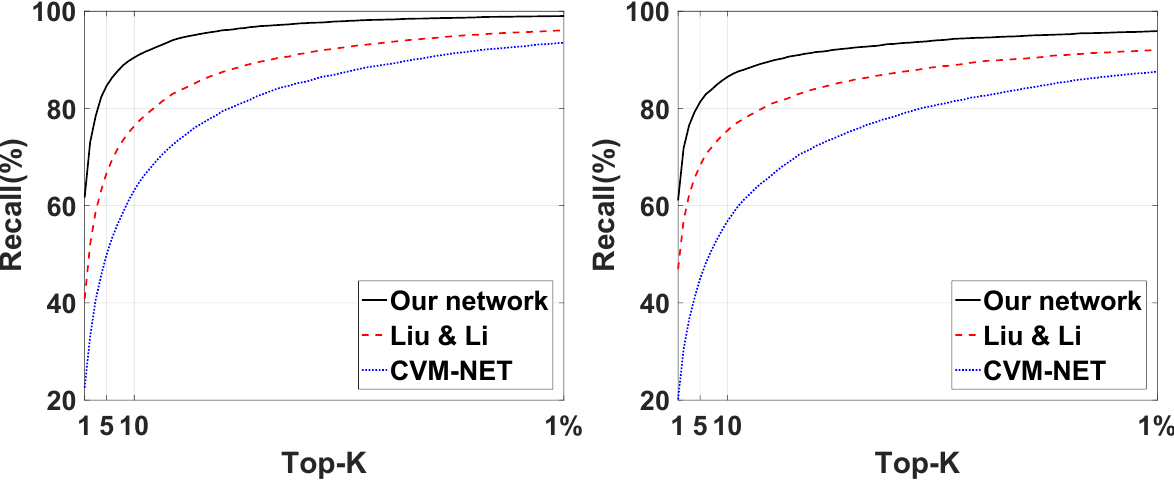}
    \caption{This figure shows that with our cross-view feature transport module, we outperform all state-of-the-art methods by a large-margin on CVUSA (\textbf{Left}) and CVACT\_val (\textbf{Right}) datasets.}
    \label{fig:ourVSstoa}
\end{figure}

\begin{table}[!t]{{ 
\setlength{\tabcolsep}{2.5pt}
\caption{{Comparison of r@1\% by state-of-the-art methods on CVUSA dataset.}}

\begin{tabular}{c|c|c|c|c|c|c}
\hline 
 & Workman  & Zhai & Vo  & CVM-NET & Liu  & Ours \\ 
 \hline
r@1\%    & 34.30 & 43.20   & 63.70 & 93.62 & 96.12 & \textbf{99.02} \\ 
\hline
\end{tabular}
\label{tab::top1_per}}} 
\end{table}

\subsection{Comparisons with the State-of-the-Art}
\subsubsection{Retrieval on Cross-View Geo-localization:}
We compare our method with the state-of-the-art cross-view localization methods, including Workman \etal \cite{workman2015wide}, Vo \etal \cite{vo2016localizing}, Zhai \etal \cite{zhai2017predicting}, CVM-NET \cite{Hu_2018_CVPR} and Liu \& Li \cite{Liu_2019_CVPR}. Following the previous works, we also report the recall at top 1\% performance in Table \ref{tab::top1_per}. 

As indicated by Table \ref{tab::top1_per}, our method outperforms the state-of-the-art methods significantly. However, as the size of the database increases, the number of gallery images increases. 
Thus, using the retrieved results at top 1\% may be impractical. 
For instance, there would be more than 80 images in the retrieved results in CVUSA.
Therefore, using recalls at top-1, 5, 10 would be more reasonable since the results can be easily analyzed by humans.
The first two rows in Table \ref{tab::recall_compare_baseline} illustrate the performance of state-of-the-art methods on recalls at top-1, 5, 10, and Figure \ref{fig:ourVSstoa} presents the complete r@K performance. Compared to the state-of-the-art, our method outperforms the second best method \cite{Liu_2019_CVPR} by a large margin. In particular, our network improves $20.64\%$ and $14.09 \%$ on the recalls at top-1 for CVUSA and CVACT\_val datasets, respectively. Table \ref{tab::recall_compare_baseline} also indicates our method is more practical for real-world cases.

\subsubsection{Accurate Geo-localization:}
We also conduct an evaluation on the large-scale CVACT\_test dataset \cite{Liu_2019_CVPR} to show the effectiveness of our method for accurate city-scale Geo-localization applications. In addition, we also compare with the state-of-the-art methods, CVM-NET \cite{Hu_2018_CVPR} and Liu \& Li \cite{Liu_2019_CVPR}, on this task.

\begin{figure}[!t]
    \centering
    \includegraphics[width=0.55\columnwidth]{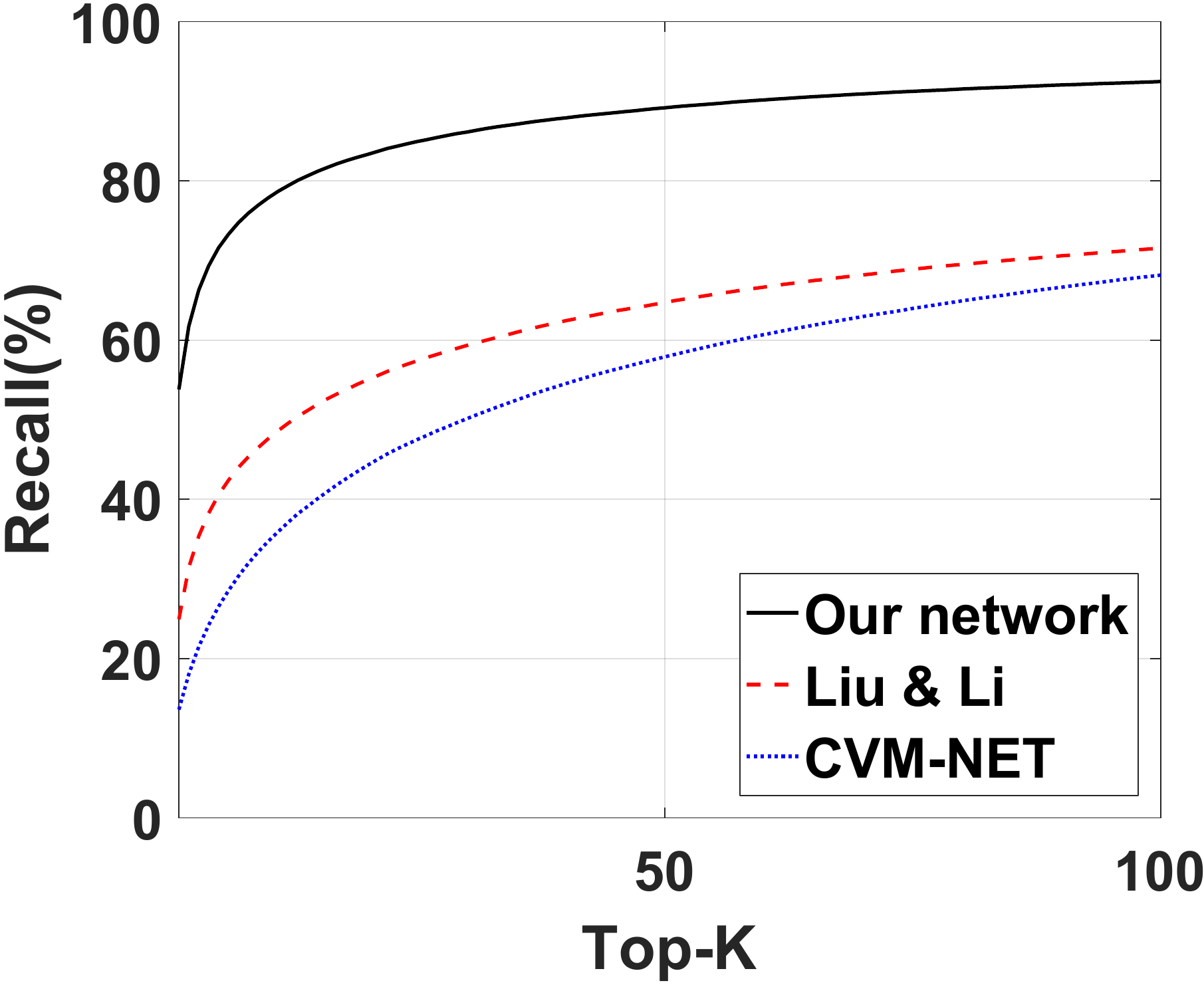}
    \caption{This graph shows that we obtain much higher localization performance (measured by recalls) than previous state-of-the-art methods.}
    \label{fig:ourGeoLoca}
\end{figure}

\begin{figure}[!t] 
    \centering
    \includegraphics[width=\columnwidth]{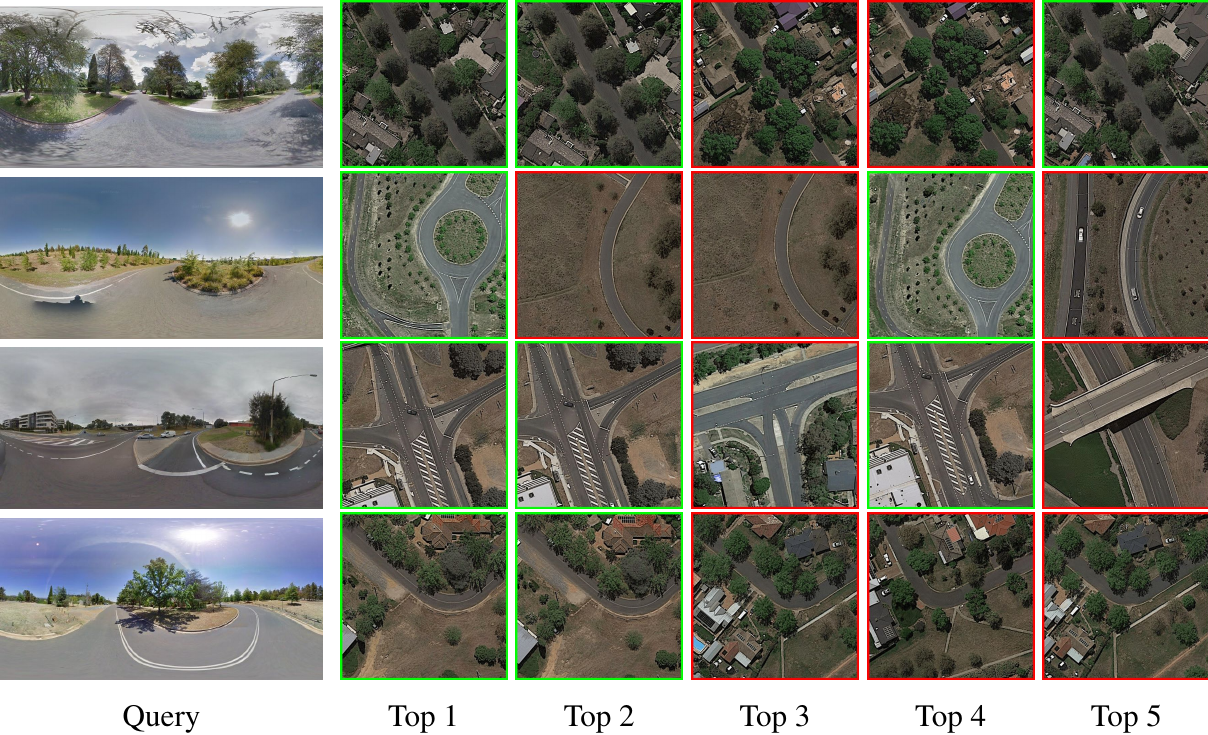}
    \caption{Sample localization results by our method on ACT\_test dataset. From left to right: ground-view query image and the Top 1-5 retrieved satellite images. Green and red borders indicate correct and incorrect retrieved results, respectively. Note that our method can localize versatile query images (\eg, suburbs and traffic intersection). (Best viewed in color on screen.)}
    \label{fig:ACT_test_local_examp}
\end{figure}

\begin{figure}[!t] 
    \centering
    \includegraphics[width=\columnwidth]{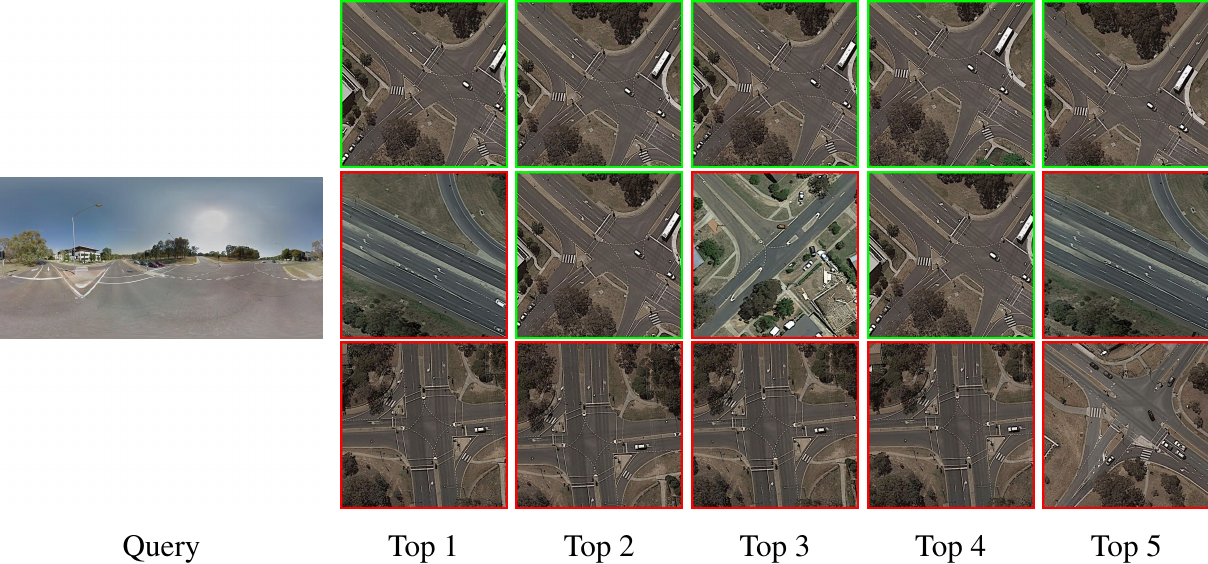}
    \caption{Comparisons of different methods on ACT\_test dataset. first row: top-5 recall results by our method; Second row: Liu \& Li result \cite{Liu_2019_CVPR}; third row: CVM-NET \cite{Hu_2018_CVPR}. Green and red borders indicate correct and incorrect retrieved results, respectively. Note that our method remains robust for this challenging scenario with many repetitive features.}
    \label{fig:ACT_test_compare}
\end{figure}

In this experiment, we follow the evaluation protocol used in \cite{arandjelovic2016netvlad}. To be specific, a query image is deemed correctly geo-localized if at least one of the top $K$ retrieved database satellite images is within the distance $d = 25m$ from the ground-truth position of the query image.  The percentage of correctly localized queries (recall) is then plotted for different values of $K$, marked as recall@$K$. The r@K performance is shown in Figure \ref{fig:ourGeoLoca}. As expected, our method still significantly outperforms the second-best method \cite{Liu_2019_CVPR}, with an improvement of $28.87\%$ at top-1. We also show some localization examples in Figure \ref{fig:ACT_test_local_examp} and \ref{fig:ACT_test_compare}, thus demonstrating the superiority of our method for localizing cross-view images in different scenarios.

\subsection{Robustness to Orientation Perturbations} The ground-level panoramas and overhead-view aerial images in the CVUSA and CVACT datasets are north aligned. 
However, it is worthy investigating the impact of orientation perturbation on the cross-view localization performance.
As the north direction of a satellite map is often available, we discuss the case that the north direction of a ground-level panorama is not given. 
In practice, we can estimate an approximate north direction by exploiting some scene clues, such as shadows of trees. 
Thus we examine the performance of our algorithm on $\pm20^\circ$ orientation noise. 
The recall results are presented in the last row in Table \ref{tab::recall_compare_baseline}.
It can be seen that our method still outperforms the state-of-the-art (CVM-NET and Liu \& Li) although they are tested on the orientation aligned ground-level panoramas.  

\section{Conclusion}
We propose a new cross-view image-based geo-localization deep network in this paper.  Our major contribution as well as the central innovation is the introducing of a novel feature transport module (CVFT) which effectively bridges the cross-view domain gap by transforming features from one domain to the other. 
In contrast to conventional approaches using deep metric learning to match corresponding feature pairs, our method provides a superior and sensible solution. We believe the key idea of this paper (explicitly handling domain differences) is also valuable for solving many other problems in computer vision and machine learning. For the task of large-scale cross-view localization, our method significantly improves performance in terms of top-K recall rate, demonstrating the power of our feature transport framework.  
Our method currently assumes the input query image is a full 360$^\circ$ panorama, 
but we argue this restriction can be relaxed under the same theoretical framework of CVFT, and this is left as a possible future extension.

\section{ Acknowledgments}
This research is supported in part by the Australia Research Council ARC Centre of Excellence for Robotics Vision (CE140100016),  ARC-Discovery (DP 190102261) and ARC-LIEF (190100080), and in part by a research gift from Baidu RAL (ApolloScapes-Robotics and Autonomous Driving Lab). The first author is a China Scholarship Council (CSC)-funded PhD student to ANU. We gratefully acknowledge the GPU gift donated by NVIDIA Corporation. We thank all anonymous reviewers for their constructive comments.

\bibliography{Bibliography-File}
\bibliographystyle{aaai}

\newpage
\onecolumn
\appendix

\section{Additional Visualization of Cross-view Feature Alignment}
\begin{figure*}[!ht]
    \centering
    
    \begin{subfigure}[t]{0.49\textwidth}
    \centering
    \includegraphics[width=0.64\linewidth,height=0.32\linewidth]{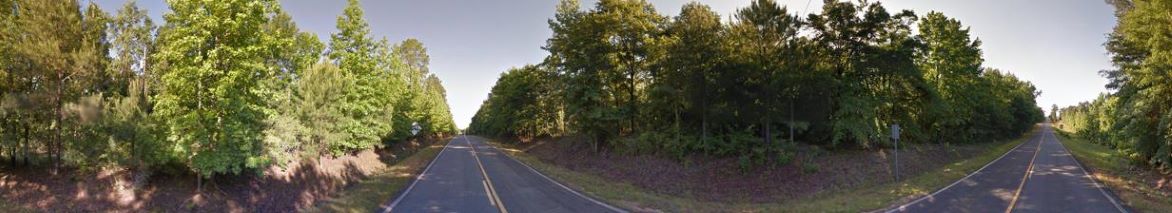}
    \includegraphics[width=0.32\linewidth]{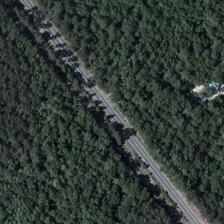}
    \includegraphics[width=0.32\linewidth]{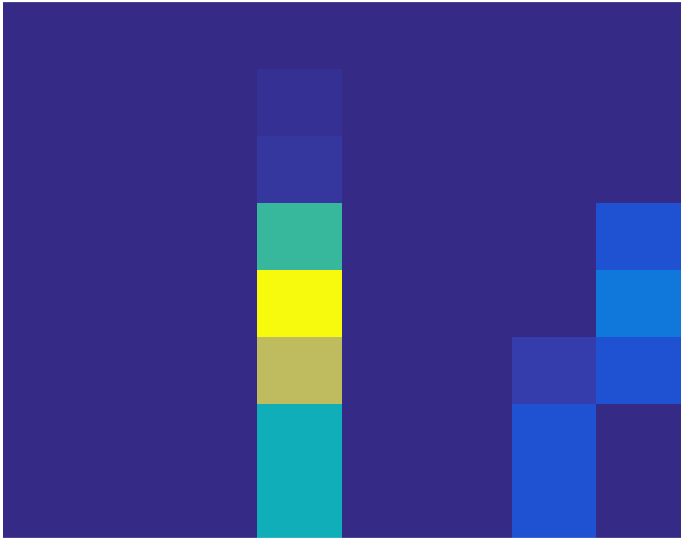}
    \hspace{-0.35em}
    \includegraphics[width=0.32\linewidth]{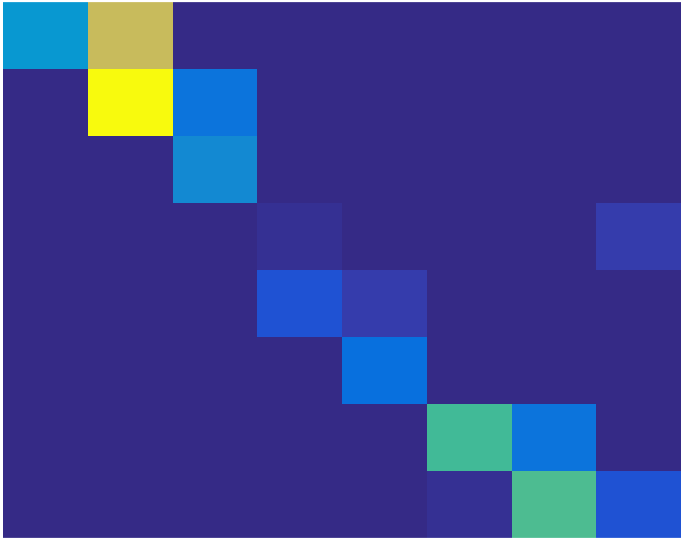}
    \hspace{-0.35em}
    \includegraphics[width=0.32\linewidth]{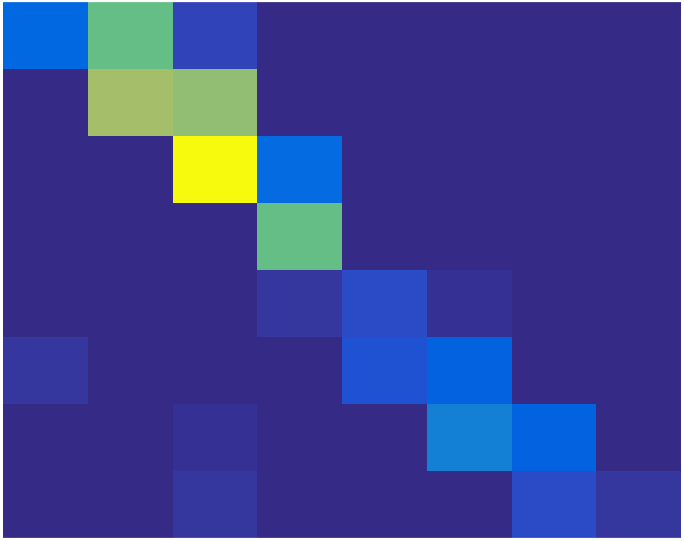}
    {\caption{}}
    \end{subfigure}
    \hspace{-0.5em}
    \begin{subfigure}[t]{0.49\textwidth}
    \centering
    \includegraphics[width=0.64\linewidth,height=0.32\linewidth]{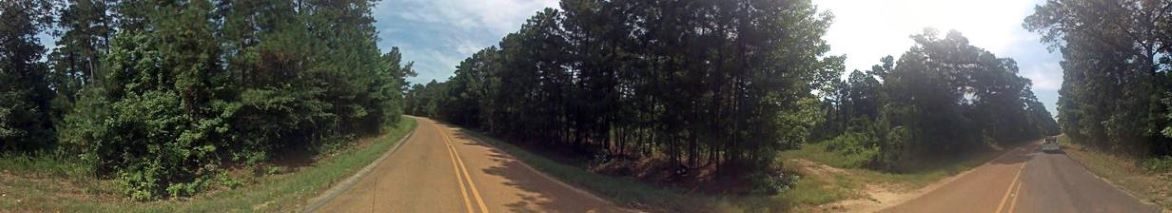}
    \includegraphics[width=0.32\linewidth]{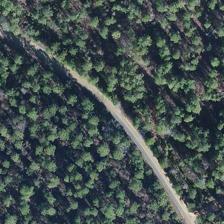}
    \includegraphics[width=0.32\linewidth]{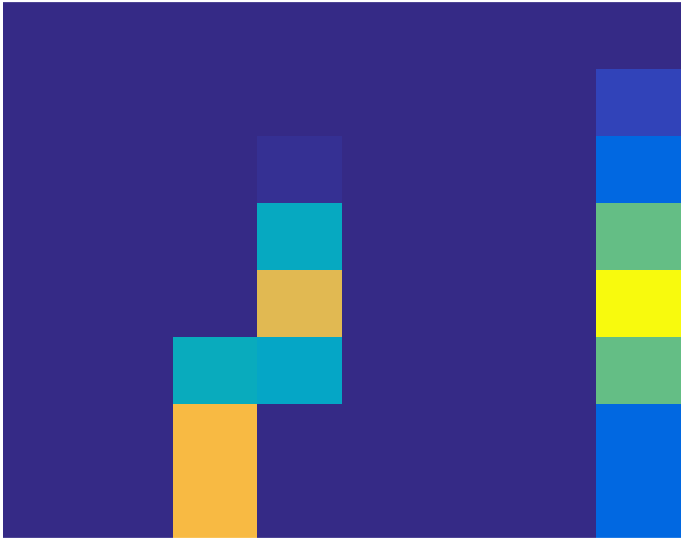}
    \hspace{-0.35em}
    \includegraphics[width=0.32\linewidth]{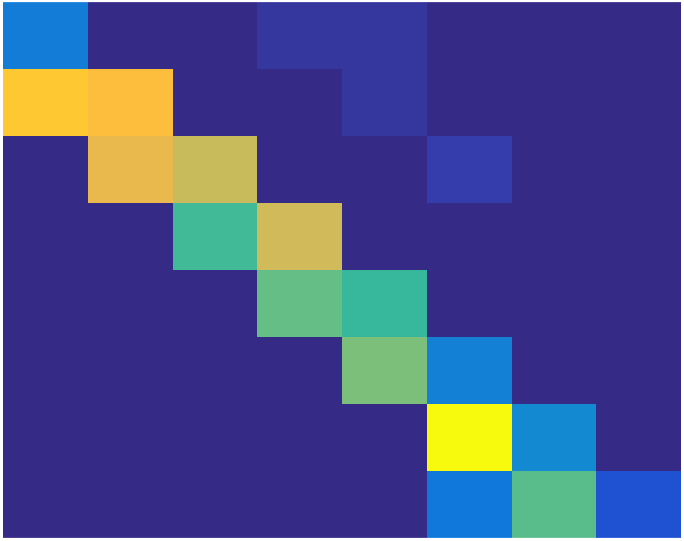}
    \hspace{-0.35em}
    \includegraphics[width=0.32\linewidth]{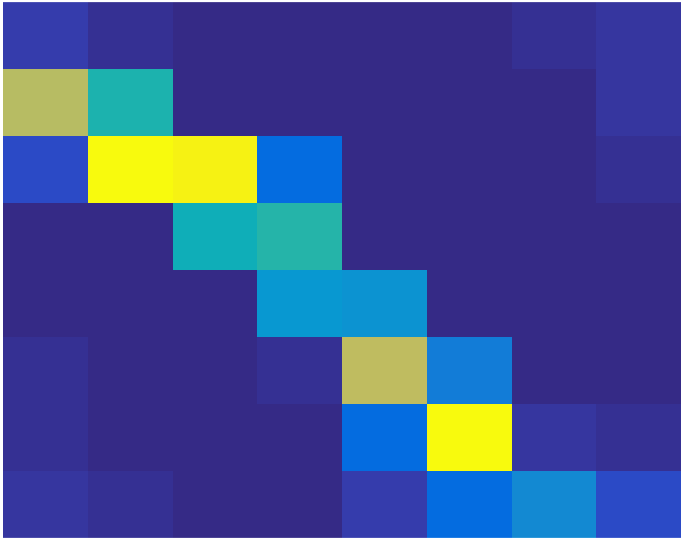}
    {\caption{}}
    \end{subfigure}
    \begin{subfigure}[t]{0.49\textwidth}
    \centering
    \includegraphics[width=0.64\linewidth,height=0.32\linewidth]{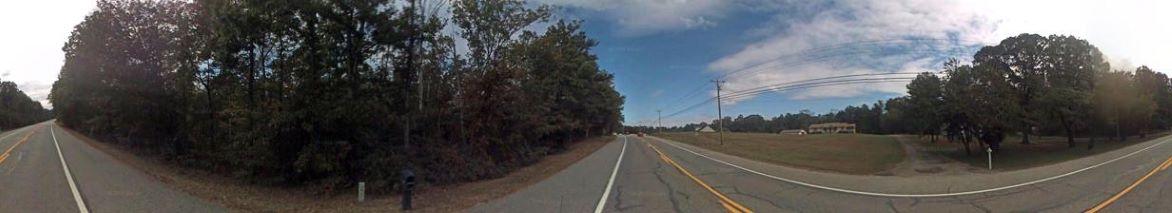}
    \includegraphics[width=0.32\linewidth]{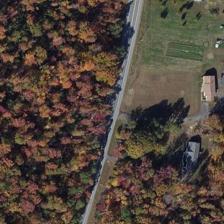}
    \includegraphics[width=0.32\linewidth]{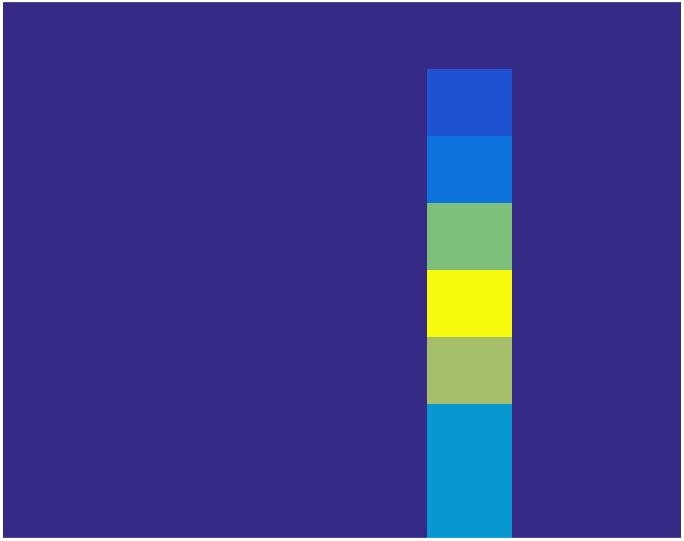}
    \hspace{-0.35em}
    \includegraphics[width=0.32\linewidth]{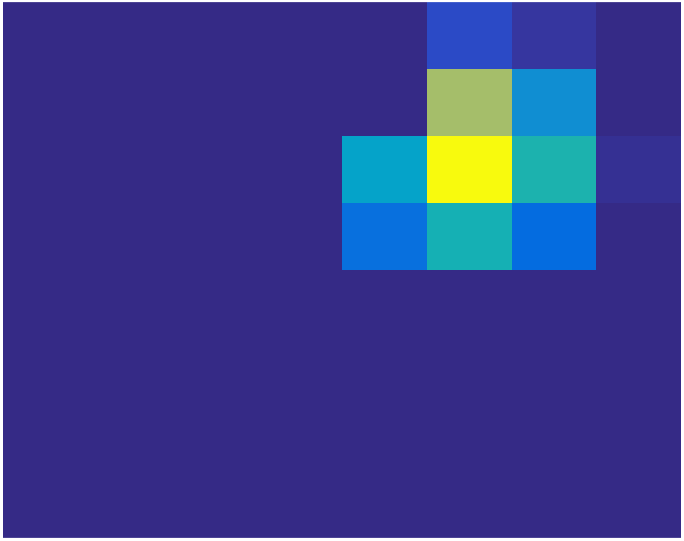}
    \hspace{-0.35em}
    \includegraphics[width=0.32\linewidth]{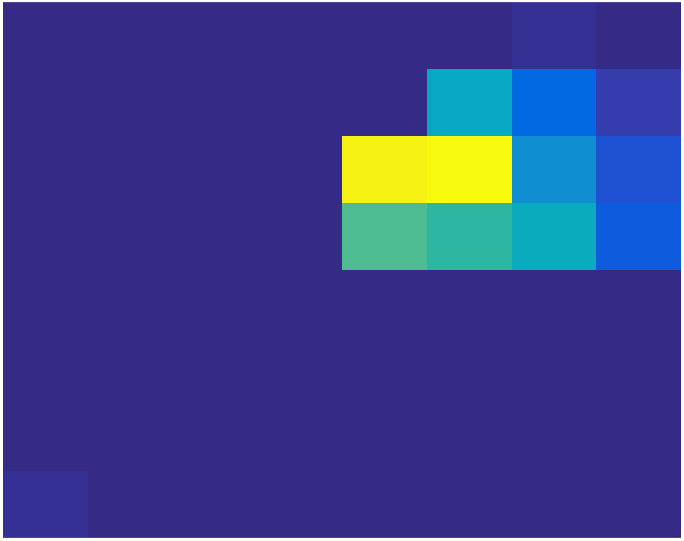}
    {\caption{}}
    \end{subfigure}
    \hspace{-0.5em}
    \begin{subfigure}[t]{0.49\textwidth}
    \centering
    \includegraphics[width=0.64\linewidth,height=0.32\linewidth]{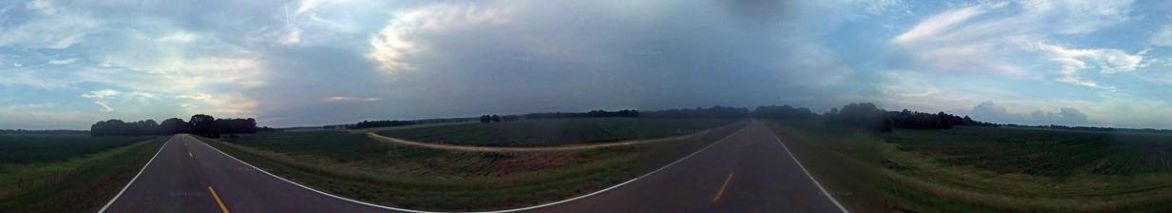}
    \includegraphics[width=0.32\linewidth]{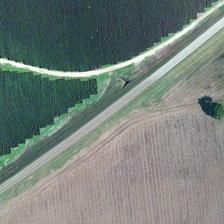}
    \includegraphics[width=0.32\linewidth]{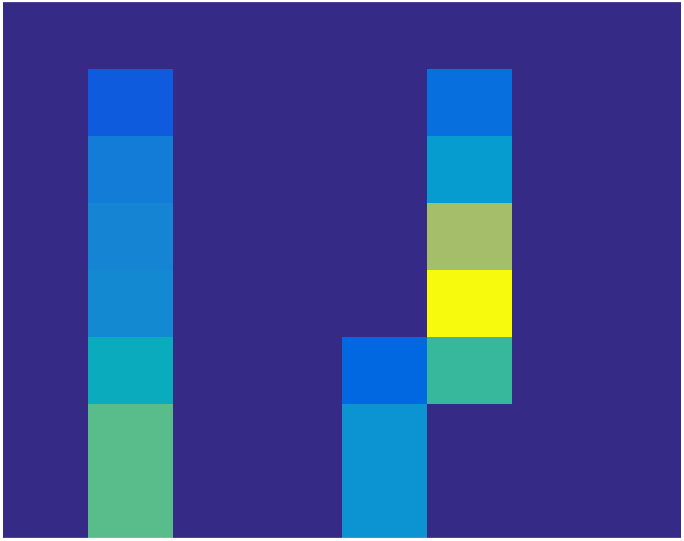}
    \hspace{-0.35em}
    \includegraphics[width=0.32\linewidth]{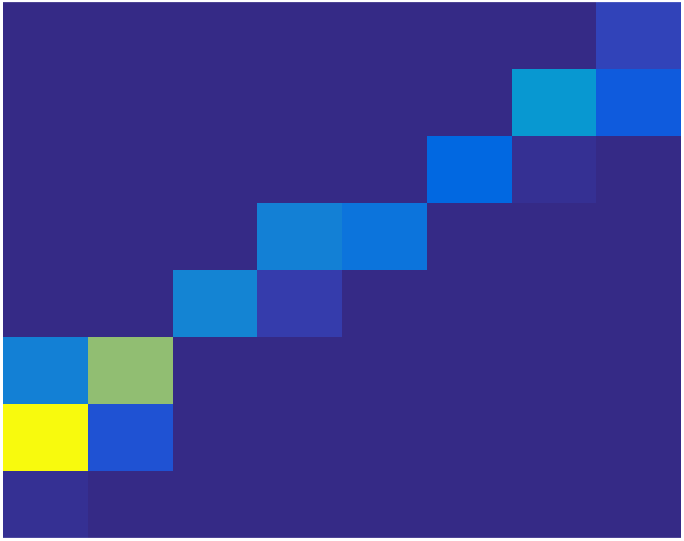}
    \hspace{-0.35em}
    \includegraphics[width=0.32\linewidth]{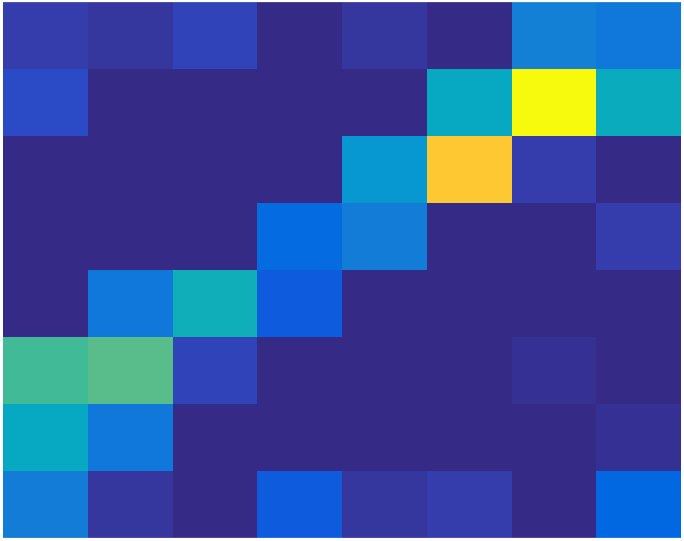}
    {\caption{}}
    \end{subfigure}
    \begin{subfigure}[t]{0.49\textwidth}
    \centering
    \includegraphics[width=0.64\linewidth,height=0.32\linewidth]{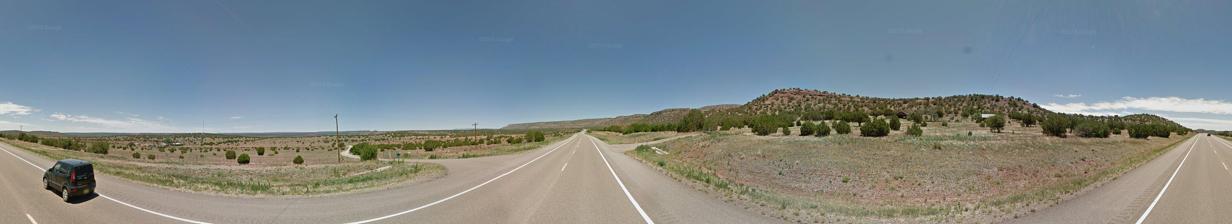}
    \includegraphics[width=0.32\linewidth]{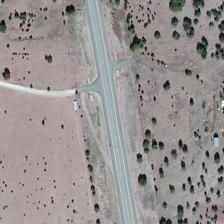}
    \includegraphics[width=0.32\linewidth]{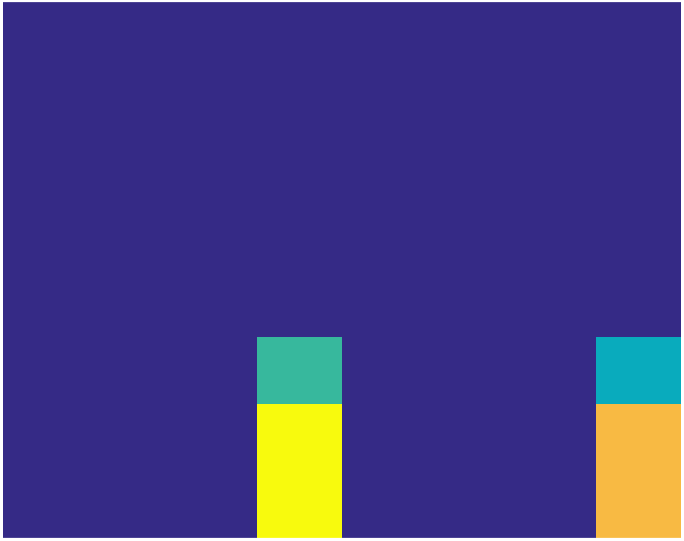}
    \hspace{-0.35em}
    \includegraphics[width=0.32\linewidth]{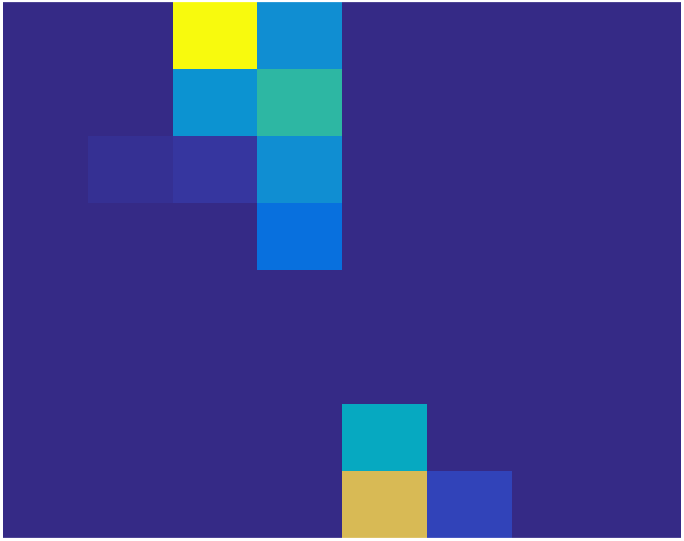}
    \hspace{-0.35em}
    \includegraphics[width=0.32\linewidth]{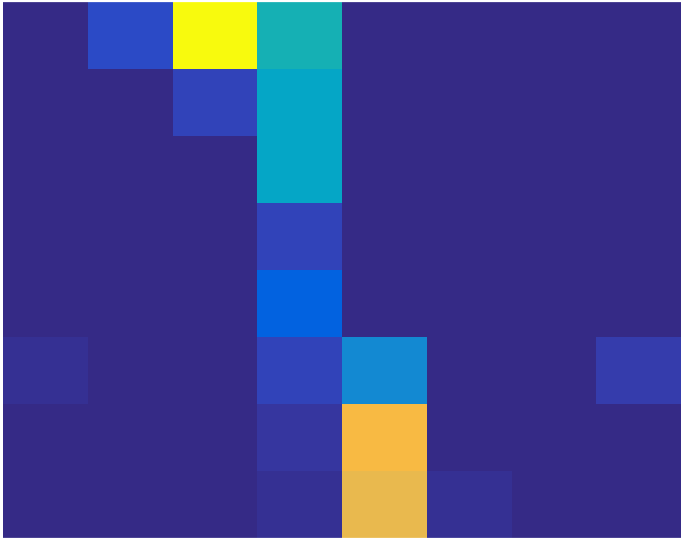}
    {\caption{}}
    \end{subfigure}
    \hspace{-0.5em}
    \begin{subfigure}[t]{0.49\textwidth}
    \centering
    \includegraphics[width=0.64\linewidth,height=0.32\linewidth]{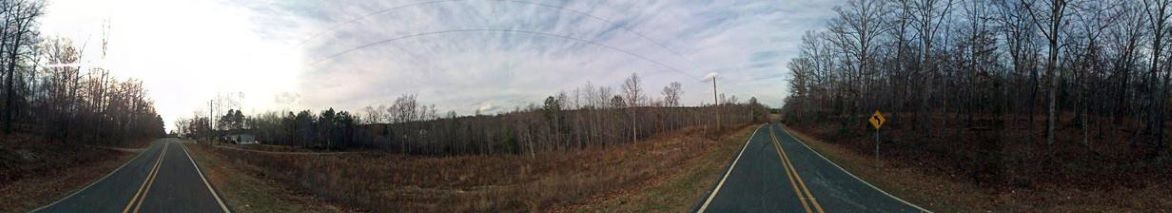}
    \includegraphics[width=0.32\linewidth]{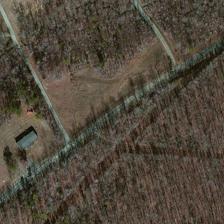}
    \includegraphics[width=0.32\linewidth]{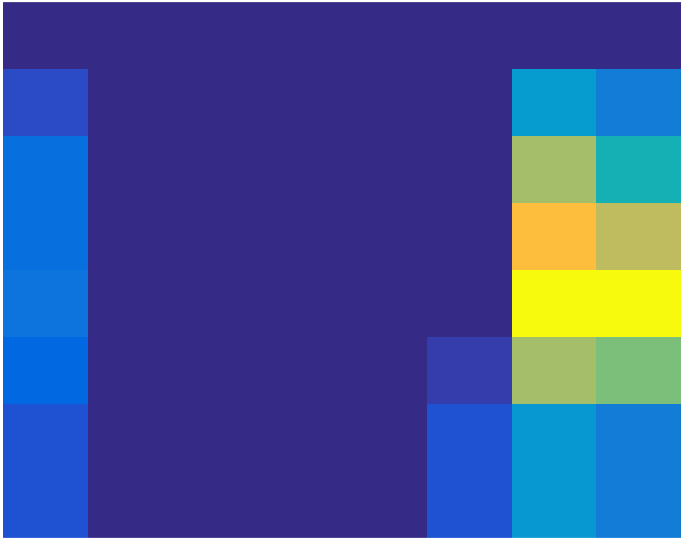}
    \hspace{-0.35em}
    \includegraphics[width=0.32\linewidth]{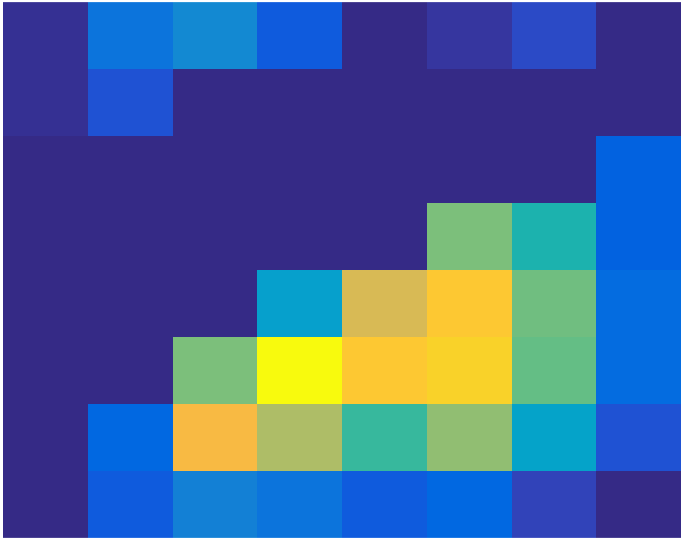}
    \hspace{-0.35em}
    \includegraphics[width=0.32\linewidth]{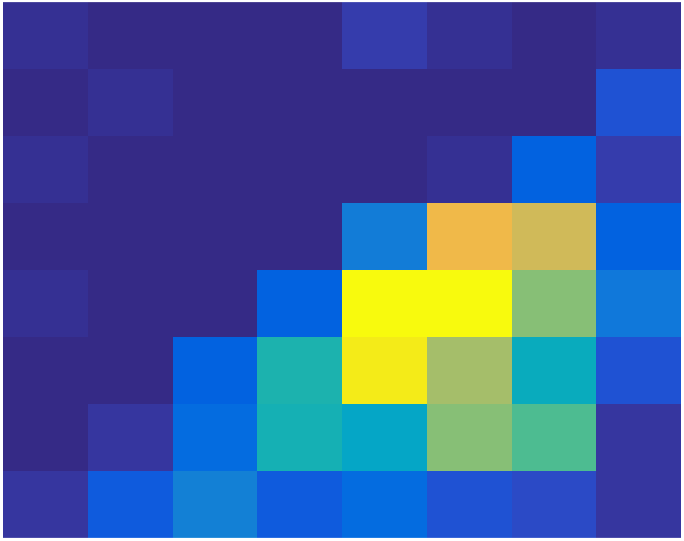}
    {\caption{}}
    \end{subfigure}
 
    \caption{Additional examples for demonstrating the effectiveness of the proposed CVFT on aligning cross-view feature maps. In the second row of each subfigure, (Left column:) Input ground-view feature map; (Middle column:) the corresponding satellite image feature map; (Right column:) CVFT-transported ground-view feature map. Note the CVFT-transported ground-view feature map and original satellite image feature map are well-aligned.}
    \label{fig:additional_visualize}
\end{figure*}

Additional examples of cross-view feature maps before and after our cross-view feature transport (CVFT) module are presented in Figure \ref{fig:additional_visualize}. It demonstrates that the proposed CVFT is able to align the feature maps of the cross-view images effectively. 

\section{Comparison with STN}
In the cross-view image based geo-localization, the ground-view panoramas capture scenes in 360 degrees while being mapped to images via the equirectangular projection, and the satellite-view images are captured perpendicular to the ground plane. Therefore, these two domain images share the same contents but exhibiting large appearance differences. 

Conventional STNs mainly regress geometric transforms of several parameters (\eg, affine, or thin plate spline). However, the cross-view transformation cannot be explictly formulated by such geometric transform. Thus, the standard STNs are unable to learn the cross-view transformations. 
To support our arguments, we carry out one experiment of using STNs to establish the cross-view transform.
To be specific, we replace the CVFT module in our framework with STNs using 2D-affine and thin plate spline transformations, denote as STN\_affine and STN\_tps, respectively. The recall results are presented in Table \ref{tab::recall_compare_stn}. For better comparison, we also present the recall results of the baseline network (Our network wo/ CVFT) and our proposed network in Table \ref{tab::recall_compare_stn}. 

As indicated in Table \ref{tab::recall_compare_stn}, using 2D-affine or thin plate spline transform based STNs, the performance of the networks is inferior to the performance of our proposed network. 
It further implies that the relationship between the cross-view image features cannot be formulated by simple geometric transforms with a few parameters. 

\section{The Impact of Dimensionality Reduction}
The dimension of our feature embedding is $4096$, which is as the same as that of CVM-net \cite{Hu_2018_CVPR}. In Liu \etal 's work \cite{Liu_2019_CVPR}, the dimension is $1536$. To illustrate the impact of feature embedding dimensions on cross-view localization performance, we resize the output feature map to different spatial sizes, thus leading to different feature embedding dimensions. The recall performance of the proposed CVFT in different feature embedding dimensions on CVUSA \cite{zhai2017predicting} and CVACT \cite{Liu_2019_CVPR} datasets is given in Table \ref{tab::recall_compare_dimension}. It shows that our  recall performance decreases gradually with the decrease of feature embedding dimension, yet still outperforms state-of-the-art methods \cite{Hu_2018_CVPR,Liu_2019_CVPR}. 

\section{Visualization of Cross-view Embeddings}
\begin{figure*}[!ht]
    \centering
    \includegraphics[width=0.8\linewidth]{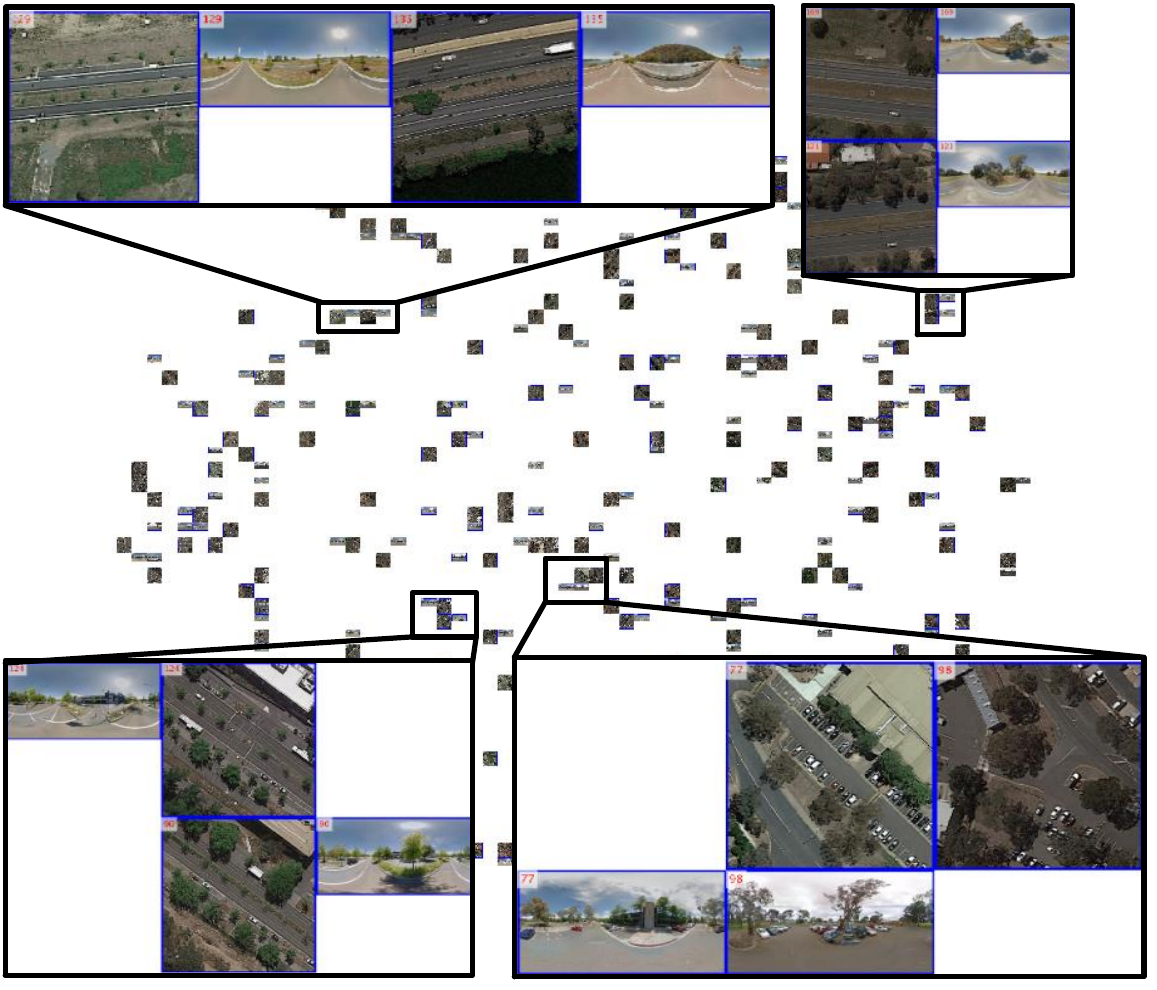}
    \caption{ Visualization of cross-view feature embeddings using t-SNE \cite{maaten2008visualizing} on CVACT \cite{Liu_2019_CVPR} dataset. The red number on the left-top corner of each image is the ID of a cross-view image pair. (Best viewed on screen with zoom-in.)}
    \label{fig:t-SNE}
\end{figure*}

The goal of our method is to embed cross-view image features under a same space, where cross-view features of the same scene lie close to each other. 
To verify it, we adopt t-SNE \cite{maaten2008visualizing} to visualize the cross-view embeddings of CVACT \cite{Liu_2019_CVPR} dataset by our method. The results as seen in Figure \ref{fig:t-SNE} illustrate that the matched cross-view image pairs are embedded into nearby positions in the embedding space despite significant view point changes.

\begin{table*}[!ht]
\setlength{\tabcolsep}{8.5pt}
\centering
\caption{Recall performance comparison between our framework and STNs.}
\label{tab::recall_compare_stn}
\begin{tabular}{c|c|c|c|c|c|c|c|c}
\hline
\multirow{2}{*}{} & \multicolumn{4}{c|}{CVUSA} & \multicolumn{4}{c}{CVACT\_val} \\ \cline{2-9} 
                       & r@1        & r@5        & r@10     & r@1\%     & r@1       & r@5       & r@10      & r@1\%   \\ \hline \hline
Our network wo/ CVFT   & 41.68      & 70.71      & 80.71    & 97.70         & 45.28     & 71.74     & 80.14     & 95.28 \\ 
STN\_affine            & 37.29      & 66.75      & 77.03    & 96.06         & 35.64     & 63.67     & 73.30     & 92.77 \\ 
STN\_tps               & 42.38      & 71.16      & 81.22    & 97.13         & 41.61     & 68.69     & 77.03     & 93.33 \\ 
Our network            &  \textbf{61.43}      & \textbf{84.69}      & \textbf{90.49}    & \textbf{99.02}   & \textbf{61.05}     & \textbf{81.33}     & \textbf{86.52} & \textbf{95.93}      \\   \hline

\end{tabular}
\end{table*}

\begin{table*}[!ht]
\setlength{\tabcolsep}{6pt}
\centering
\caption{Recall performance of our CVFT under different feature embedding  dimensions.}
\label{tab::recall_compare_dimension}
\begin{tabular}{c|c|c|c|c|c|c|c|c}
\hline
\multirow{2}{*}{} & \multicolumn{4}{c|}{CVUSA} & \multicolumn{4}{c}{CVACT\_val} \\ \cline{2-9} 
       & r@1        & r@5        & r@10     & r@1\%     & r@1       & r@5       & r@10      & r@1\%   \\ \hline\hline
4096   & 61.43      & 84.69      & 90.49    & 99.02         & 61.05     & 81.34     & 86.52     & 95.93 \\ 

2048   & 56.43      & 80.77      & 88.07    & 98.81         & 57.86     & 79.69     & 85.25     & 95.49 \\ 
1536   & 52.22      & 77.30      & 85.31    & 98.33         & 53.94     & 77.40     & 83.27     & 95.09 \\ 
512    & 43.05      & 69.42      & 78.66    & 96.89         & 44.69     & 70.19     & 77.81     & 93.56 \\ 
\hline
4096 (CVM-NET \cite{Hu_2018_CVPR})               & 22.47      & 49.98      & 63.18     & 93.62     & 17.53     & 42.13 & 53.15     & 86.72 \\ 
1536 (Liu \& Li \cite{Liu_2019_CVPR})               & 40.79      & 66.82      & 76.36     & 96.12     & 46.96     & 68.28 & 75.48     & 92.04 \\ \hline
\end{tabular}
\end{table*}

\end{document}